# MOAR Planner: Multi-Objective and Adaptive Risk-Aware Path Planning for Infrastructure Inspection with a UAV

Louis Petit[1] and Alexis Lussier Desbiens[1]

***Abstract*— The problem of autonomous navigation for UAV inspection remains challenging as it requires effectively navigating in close proximity to obstacles, while accounting for dynamic risk factors such as weather conditions, communication reliability, and battery autonomy. This paper introduces the MOAR path planner which addresses the complexities of evolving risks during missions. It offers real-time trajectory adaptation while concurrently optimizing safety, time, and energy. The planner employs a risk-aware cost function that integrates precomputed cost maps, the new concepts of damage and insertion costs, and an adaptive speed planning framework. With that, the optimal path is searched in a graph using a discrete representation of the state and action spaces. The method is evaluated through simulations and real-world flight tests. The results show the capability to generate real-time trajectories spanning a broad range of evaluation metrics—around 90% of the range occupied by popular algorithms. The proposed framework contributes by enabling UAVs to navigate more autonomously and reliably in critical missions.**

## I. INTRODUCTION

Autonomous robot navigation is a major field of research with applications in areas such as search and rescue [1], surveillance [2] and inspection [3]. In this work, we consider the problem of path planning with an unmanned aerial vehicle (UAV) for inspection. The main objective of such missions is to effectively navigate around and in close proximity to the obstacles that require assessment. These missions often poses challenges for path planning, e.g., battery management, real-time computation, collision and damage risks near costly assets, and optimizing mission duration to capitalize on favorable weather conditions. Some challenges also arise and evolve during the mission due to risks such as sporadic sensor failure or gusty winds. This changing and challenging environment makes for careful flight planning to ensure mission success.

Despite numerous studies on Multi-Objective Path Planning (MOPP), there is currently no method, to our knowledge, that offers real-time trajectory adaptation to evolving mission risks [4]–[8]. This paper introduces a novel algorithm for Multi-Objective and Adaptive Risk-aware (MOAR) path planning. MOAR optimizes trajectories for inspection missions, considering safety, time, and energy objectives. It dynamically adjusts trajectories in response to mission risks, such as weather conditions, remaining flight autonomy, and the probability of communication or localization loss. Using precomputed safety cost maps and a variable speed map, we find the optimal path in a 3D graph by minimizing a modulated cost function. Benchmark simulations demonstrate MOAR's flexibility, generating trajectories spanning up to 150% of benchmark algorithms' metrics (e.g., path length). Also, the risk adaptation framework enables our method to achieve around 90% coverage of desirable trajectories, i.e. from the shortest to the safest, in path length and obstacle clearance, for various risk levels.

[1]The authors are with the Createk Design Lab, University of Sherbrooke, Sherbrooke, J1K 2R1, Canada louis.petit@usherbrooke.ca; alexis.lussier.desbiens@usherbrooke.ca

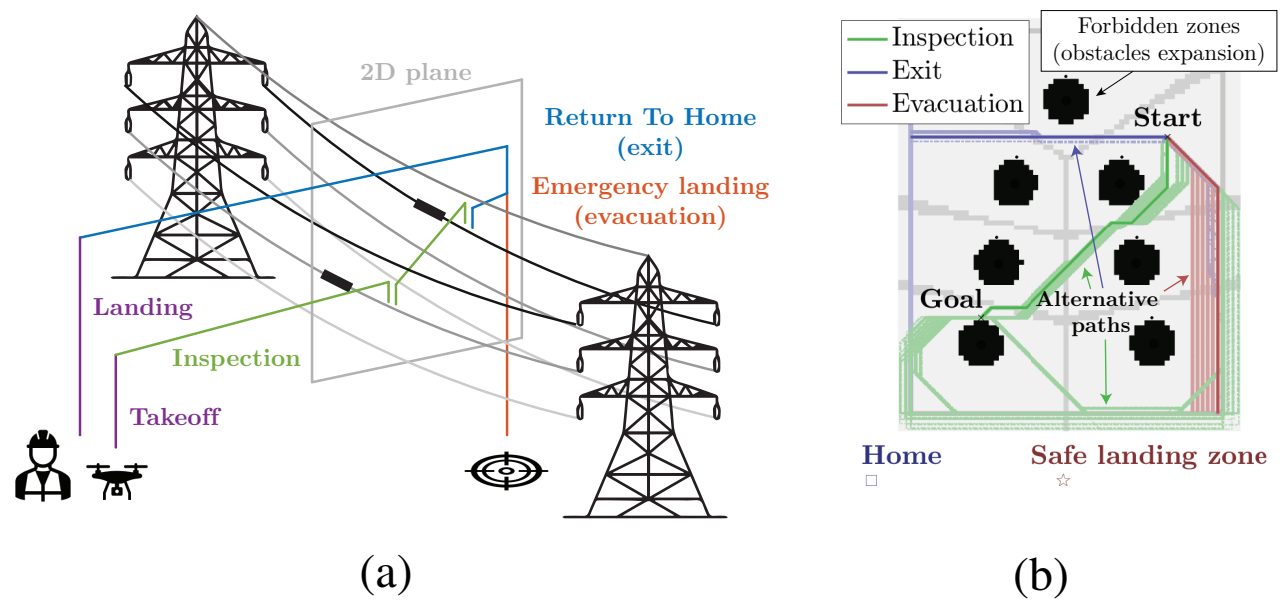

(a) (b)

Fig. 1: Example of a power line inspection mission: (a) Schematic representation of an inspection of 2 splices (black rectangles). (b) Path found with our planner (bold), and alternative paths for different levels of risk.

In this paper, we apply our method to a power line inspection problem [9], [10], using the LineDrone robot [11], [12]. This robot was developed by Hydro-Quebec to conduct non-destructive inspections of high-voltage transmission lines, by landing and rolling on them, and inspecting splices using an onboard probe. It helps assess the remaining useful life of conductors and improve asset management. It is equipped with a dual-antenna GPS for localization and a vertically mounted LiDAR for obstacles detection. Teleoperated inspection by a human on the ground is often impractical due to challenges such as parallax, loss of depth perception at distance, inaccessible terrains (rivers, bushes, mountains), and the vastness of the area to be navigated. A typical mission is shown in Fig. 1. Although developed primarily for power line inspection in 2D, our framework can be easily applied to other path planning problems in the proximity of obstacles, including most infrastructure inspection missions.

This paper is structured as follows: Section II defines the problem, Section III reviews related work in MOPP, Section IV details our approach, Section V presents performance comparisons, sensitivity analysis, and risk adaptability results, in simulation and real flight tests, and Section VI concludes by discussing potential future directions.

## II. PRELIMINARY MATERIAL

### A. Assumptions

Fig. 1a shows a typical power line inspection using the LineDrone. Since the drone is able to roll on the line up to the splice [11], the configuration space can be simplified to a 2D vertical plane. However, our approach is equally applicable to a 3D problem. Any rigid 2D object pose can be defined by a 2-parameter configuration: $(x, y, \phi)$ and the group of motions is the Euclidean space $SE(2) = \mathbb{R}^2 \times SO(1)$. An aerial robot used for infrastructure inspection (e.g. a quadcopter) is likely to move at low speed, thus keeping the roll parameter ($\phi$) close to zero. The problem can thus be considered holonomic with negligible dynamic constraints. Implementing an offline expansion to generate the collision-free space further simplifies the path planning to a point-like problem in $\mathbb{R}^2$. In this space, a path can be represented as a discrete set of nodes.

### B. Problem Statement

Let $C_{free}$ be the free configuration space, and $\Sigma$ be the set of all nontrivial paths. The MOPP problem is defined as the search for a path $\tau^*$ that minimizes the cost function $c : \Sigma_{C_{free}} \to \mathbb{R}_{\geq 0}$, which assigns a non-negative cost to all nontrivial paths in $C_{free}$. Given initial and final configurations $q_I$ and $q_G$, an optimal algorithm must find a path $\tau^* : [0, 1] \to C_{free}$ such that (a) $\tau^*(0) = q_I$ and $\tau^*(1) = q_G$, and (b) $c(\tau^*) = min_{\tau \in \Sigma_{C_{free}}} c(\tau)$, and report failure if no such path exists. The cost function is defined as $c = \sum_i k_i(\mathcal{R}) \cdot \mathcal{C}_i$, where $k_i$ is a coefficient, $\mathcal{R}$ is the set of risk factors (e.g., battery, wind, communication, localization), and $\mathcal{C}_i$ is a cost (e.g., time, safety, energy).

## III. RELATED WORK

The optimal path planning problem has been tackled in many works. However, no definition of optimality is widely accepted for application to any problem. Autonomous inspection robots are typically deployed on all types of terrain (on land, in the air, underwater, or on the water) and the types of planning objectives are diverse and application-dependent. In this section, we start by stating the planning objectives most commonly used in the literature. Next, we introduce MOPP techniques. Finally, we discuss how objectives are combined to determine the optimal path.

*a) Types of Objectives:* In many approaches, the path length is chosen as the optimization objective [13], [14]. In such scenarios, algorithms like A* or Dijkstra guarantee the optimality of the computed trajectory [15], [16]. For variable-speed systems, this objective can be refined into a shortest-time objective [7], [17], [18]. Some approaches derive an energy consumption objective [19], either through system modeling or experimental data, in windy conditions [20], [21] or sea currents [22]. The problem is solved using a modified version of A* with an appropriate cost function and heuristic. Other approaches maximize clearance from obstacles or danger zones using, e.g., a Voronoi graph [5], [8], [23].

*b) Multi-Objective Path Planning:* Some work has focused on several objectives simultaneously, e.g. path length and energy consumption [4], navigation time and collision risk [5]–[7], or other combinations of objectives including maneuverability [8], noise [24], path smoothness [25], or length of tether unwound [26].

*c) Handling of Multiple Objectives:* Some approaches perform a single-objective optimization (e.g., path length) and add constraints (e.g., safety distance) to find a sub-optimal solution that implicitly incorporates other objectives [27]. Other approaches handle multiple objectives through a weighted aggregation of objectives into a single function [6], [28]. Other works find Pareto optimal solutions using a multi-objective version of solvers like A* [4], the evolutionary algorithm [29] or the genetic algorithm [7], [30].

However, all these approaches provide an optimal solution to a single planning problem. During an inspection mission with an autonomous drone, the optimization objective typically varies over time and depends on several flight parameters. For instance, in windy conditions or situations with a high risk of position loss, it is desirable to increase the clearance distance. Conversely, a low battery requires faster and more energy-efficient navigation. Furthermore, these approaches do not provide a framework that can adapt trajectories spatially, e.g., by allowing the robot to navigate faster when in a safer region. Finally, existing studies only take into account the obstacles or danger zones clearance when calculating the collision cost. These studies do not propose to calculate a cost linked to the risk of damage in the event of a drone crash, or to the degree of insertion into obstacle structures. In this paper, we introduce a mission-aware adaptive algorithm designed to generate trajectories that are adjusted—in space and time—to accommodate varying risk levels inherent to the flight mission. The contributions of our method are as follows: (a) we develop a novel framework for modulating the cost function based on risk factors, (b) we define a new cost function and approach to cost integration, (c) we introduce the notion of damage cost in the event of a crash and insertion cost in the obstacle zone, (d) we use a new framework for adaptive speeds planning, (e) we provide a new representation of the state and action spaces in a discrete graph using motion primitives.

## IV. METHODOLOGY

This section starts by stating a typical mission and defining the cost function used to solve the multi-objective path planning problem. Then, the path search method is described. Finally, the real-time adjustment of the cost function coefficients according to risk factors is detailed.

### A. Multi-Objective Path Planning

We define a cost function

$$J = k_S\, \mathcal{SC} + k_T\, \mathcal{TC} + k_E\, \mathcal{EC}, \tag{1}$$

where $\mathcal{SC}$, $\mathcal{TC}$ and $\mathcal{EC}$ are respectively the safety cost, time cost and energy cost, and $k_S$, $k_T$ and $k_E$ are the associated coefficients.

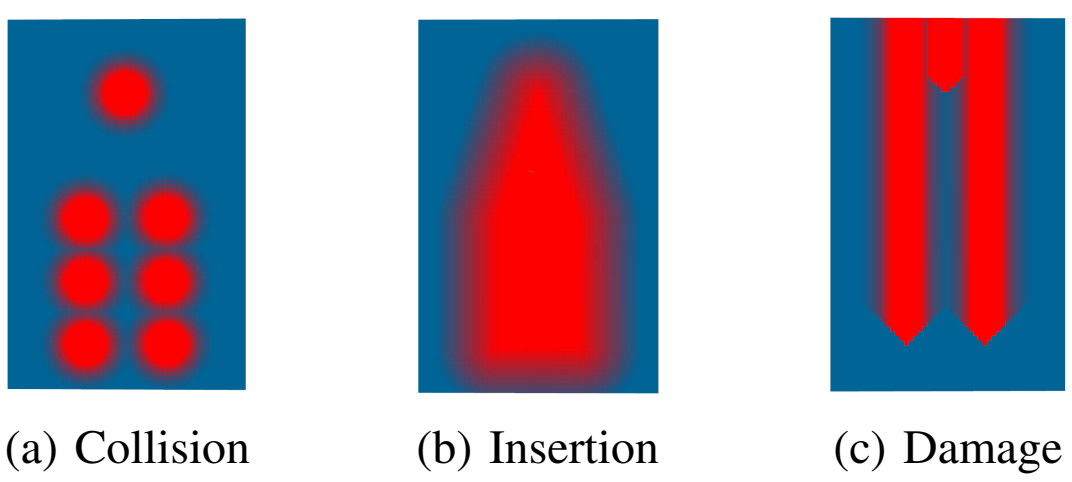

(a) Collision (b) Insertion (c) Damage

Fig. 2: Safety cost grids with $d_{min} = 1.2m$ and $d_{max} = 3m$. Red represents a cost of 1 and blue a cost of 0.

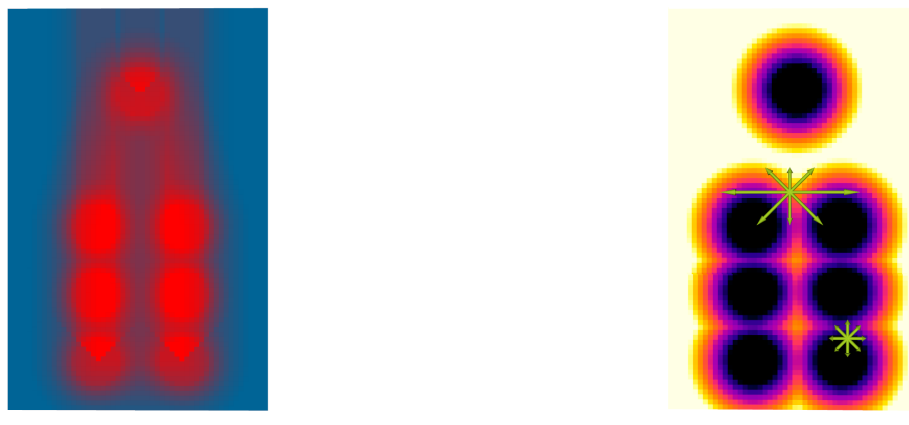

Fig. 3: Assembled safety cost grid ($\mathcal{SC}_i$) with $k_a = 0.3$, $k_b = 0.2$, and $k_c = 0.5$ (arbitrary). Red is 1 and blue is 0.

Fig. 4: Variable speed grid ($v_{sat}(d_{obs})$). Low is in black and high in yellow. Green arrows indicate $v(d_{obs}, \theta)$.

1) *Safety Cost:* The safety cost for a node $i$ in the path is composed of 3 terms (Fig. 2):

$$\mathcal{SC}_i = k_a\, C(d_{obs}) + k_b\, C(d_{ch}) + k_c\, C(d_{hor}). \tag{2}$$

The first term is a collision cost that helps maintain a safe minimum distance ($d_{obs}$) from obstacles. The second term depends on the minimum distance ($d_{ch}$) to a convex hull containing the obstacles, to favor trajectories outside the inspected structure. The third term is a damage cost that penalizes positions within a minimum horizontal distance ($d_{hor}$) above obstacles, which could lead to damage in the event of a fall. Note that the damage cost is zero if the UAV is below a 45 degree slope from the bottom obstacle. The function $C(d)$ is defined similarly to a potential field [31]:

$$C(d) = \begin{cases} 1 & \text{if } d < d_{min}, \\ \alpha(\frac{d_{max}}{d} - 1) & \text{if } d \leq d_{max}, \\ 0 & \text{if } d > d_{max}, \end{cases} \tag{3}$$

where $\alpha$ is a scaling factor equal to $\frac{d_{min}}{d_{max} - d_{min}}$. The coefficients $k_a$, $k_b$ and $k_c$ in Eq. 2 are normalized to ensure that $0 \leq \mathcal{SC}_i \leq 1$. Fig. 2 and 3 show the pre-computed safety cost maps for a power line configuration similar to Fig. 1a.

2) *Time Cost:* The time cost is computed as:

$$\mathcal{TC}_i = \frac{d_i}{v(d_{obs}, \theta)}, \tag{4}$$

where $d_i$ is the distance of the path segment, and the speed $v$ is variable depending on flight direction ($\theta$) and distance from obstacles ($d_{obs}$). Regarding dependence on flight direction, a drone can reach a higher speed in horizontal flight than, e.g., in ascending flight, with the same amount of energy. Moreover, since an inspection drone is likely to move slowly, it is assumed that it can reach cruising speed quickly. We can therefore define a fixed cruising speed $v_{max,\theta}$ for each direction of flight. For the sake of simplicity, speeds are only specified for orthogonal and diagonal flight directions. Regarding dependence on obstacle clearance, an inspection robot typically has the ability to operate at higher velocities when away from obstacles, as the precision requirements on localization, mapping, and control are reduced. We define a maximum speed variation function

$$v_{sat}(d_{obs}) = \begin{cases} v_{MIN} & \text{if } d < d_{min}, \\ v_{MIN} + \beta(d_{obs} - d_{min}) & \text{if } d \leq d_{max}, \\ v_{MAX} & \text{if } d > d_{max}, \end{cases} \tag{5}$$

where $\beta$ is a scaling factor equal to $\frac{v_{MAX} - v_{MIN}}{d_{max} - d_{min}}$. While the cruising speed $v_{max,\theta}$ depends on the direction of flight, the maximum reachable speed $v_{MAX}$ is usually equal to the horizontal cruising speed $v_{max,hor}$, for a UAV. The minimum speed $v_{MIN}$ is uniform across directions, as it is governed by the need for precision in navigation. Ultimately, navigation speed is computed as $v(d_{obs}, \theta) = min(v_{sat}(d_{obs}), v_{max,\theta})$. Fig. 4 shows an example with horizontal speed $v_{max,hor} = 10m/s$, and ascending speed $v_{max,up} = 1.5m/s$.

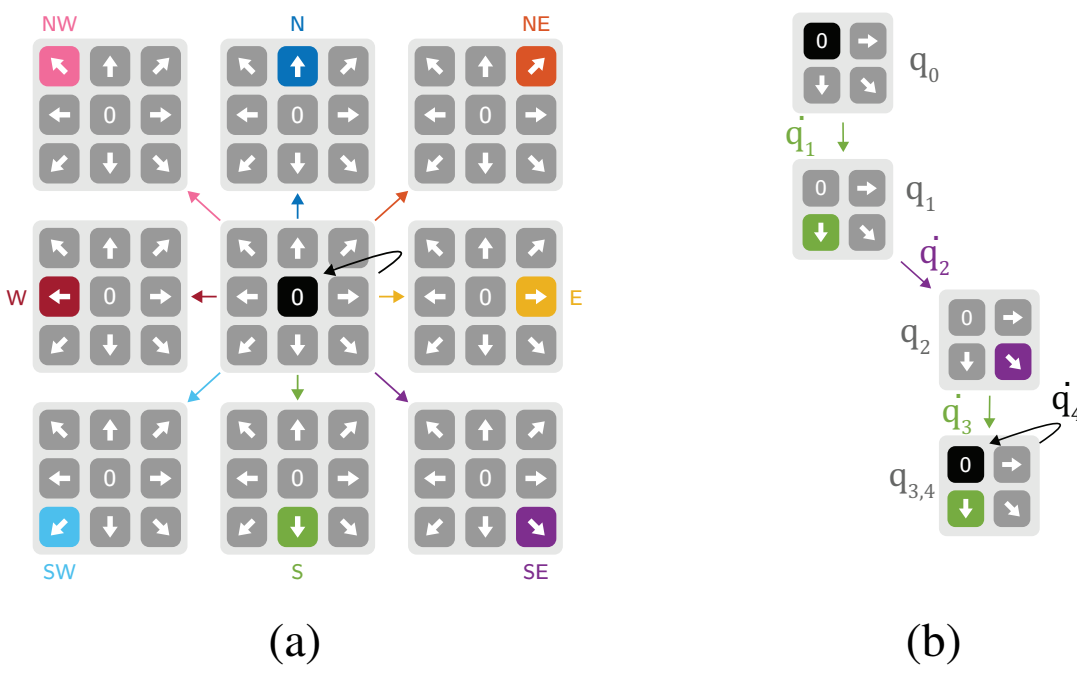

(a) (b)

Fig. 5: Representation of the graph with (a) the 9 possible actions at a given node, (b) a path example.

3) *Energy Cost:* By choosing cruising speeds adapted to the direction of travel (e.g. fast for horizontal flight, slow for ascent), we can assume that the energy consumption in a given time is similar across flight directions, in steady state. As the time cost $\mathcal{TC}$ already includes a notion of navigation time, only the transient regime is not considered in the cost function so far. Since an inspection drone is likely to be slow, we can assume that it will reach cruising speed in a similar amount of time, regardless of the velocity change. The difference in velocities is therefore a good measure of the energy consumption during acceleration:

$$\mathcal{EC}_i = \frac{\|\mathbf{v}_i - \mathbf{v}_{i-1}\|}{2\, v_{MAX}}. \tag{6}$$

Note that the energy cost is normalized with the maximum authorized velocity change to ensure $0 \leq \mathcal{EC}_i \leq 1$.

### *B. Graph Search*

We create a 3D graph of states $(x, y, \theta)$ by discretizing the 2D plane $(x, y)$ and flight directions $(\theta)$, not to be confused with robot orientation. A node in the graph connects to 9 neighboring nodes using motion primitives (Fig. 5). The set of possible actions at each node is $\{0, N, NE, E, SE, S, SW, W, NW\}$ using cardinal notation. The optimal path is then searched in the graph—excluding the occupied cells in black in Fig. 1b—using A* [15]. Pre-computed safety costs (Fig. 3) and velocities (Fig. 4) are stored in the graph to reduce redundant calculations. The cost of a path to node $n$ is computed as:

$$g_n = \sum_{i=1}^{n} k_S \, \mathcal{SC}_i \, d_i + \sum_{i=1}^{n} k_T \, \mathcal{TC}_i + \sum_{i=1}^{n} k_E \, \mathcal{EC}_i, \quad (7)$$

where $\mathcal{SC}_i$ is normalized and then integrated over the path length, $\mathcal{TC}_i$ is summed in seconds, and $\mathcal{EC}_i$ is normalized and then summed. The heuristic function from node $n$ to the goal $m$ is

$$h_n = k_T \, \frac{d(n, m)}{v_{MAX}} + k_E \, \mathcal{EC}(v_i \to 0), \quad (8)$$

which corresponds to the minimum time bound to cover the distance following diagonal or straight steps, and the acceleration cost to stop.

For exit and evacuation paths, the same method is followed, but using Dijkstra [16] instead of A* and thus omitting the heuristic in Eq. 8. The optimal path is searched to an exit or evacuation region on the grid, and the process ends when a first path has been found to a point in the goal region. This region consists of grid edges visible from the designated landing point, i.e., a straight line between the edge and landing point does not intersect the grid. Coefficients can be adjusted based on the navigation mode (inspection, exit, evacuation)—e.g., to ensure safer behavior in evacuation than in exit)—or dynamically tuned according to the current mission's risk assessment, as explained in the next section.

### C. Risk-Aware Coefficient Adjustment

We define 4 mission failure risks: (a) wind risk $\mathcal{WR}$, (b) communication risk $\mathcal{CR}$, (c) localization risk $\mathcal{LR}$, (d) battery risk $\mathcal{BR}$. These risks are empirically derived for our specific problem but can be adapted for other applications and systems. In our case, GPS signal loss, communication signal loss, and unstable windy conditions necessitate safer navigation. On the other hand, limited remaining flight autonomy calls for faster navigation. Finally, reduced autonomy and the risk of instability require smoother maneuvers, i.e. low accelerations. These risk indices are estimated between 0 and 1 in real time during the flight, based on factors such as (a) wind speed and motors saturation, (b) data transmission speed, (c) number of available GPS satellites and dilution of precision values, (d) battery level and return to home distance. We introduce 3 adjustment functions for the coefficients in Eq. 1:

$$\begin{aligned} k_S &= \gamma \, k_{S,0} \, (1 + (\frac{1}{2} \, \mathcal{WR} + \frac{1}{4} \, \mathcal{CR} + \frac{1}{4} \, \mathcal{LR} - \mathcal{BR})), \\ k_T &= \gamma \, k_{T,0} \, (1 - (\frac{1}{2} \, \mathcal{WR} + \frac{1}{4} \, \mathcal{CR} + \frac{1}{4} \, \mathcal{LR} - \mathcal{BR})), \\ k_E &= \gamma \, k_{E,0} \, (1 + (\frac{1}{2} \, \mathcal{WR} + \frac{1}{2} \, \mathcal{BR})), \\ 1 &= k_S + k_T + k_E. \end{aligned} \quad (9)$$

Here, $\gamma$ serves as a scaling factor for normalization.

## V. EXPERIMENTS

This section first presents a simulated performance comparison with popular algorithms and a parametric sensitivity analysis to the cost function coefficients. Next, we validate performance in terms of adaptability to risk levels. Finally, overall behavior is validated by real flight tests.

### A. Performance Comparison & Parametric Sensitivity

This subsection begins by detailing the simulation setup, as well as validating the solution's real-time capabilities. Next, the performance of our algorithm is compared qualitatively and then quantitatively with popular algorithms, along with a parametric sensitivity analysis.

Simulations were performed with a 2.3GHz i7 CPU running an 8-core processor and 16 GB RAM. The simulated drone is equipped with a Velodyne LiDAR oriented in the vertical plane. Obstacles are detected using the LiDAR point cloud and a segmentation algorithm based on RANSAC [32]. The vertical 2D plane is then determined perpendicular to the lines, and the conductors are represented punctually in this plane. The approach was deployed on ROS (Robot Operating System) and tested on 3 simulated environments in Gazebo. They were reconstructed from official measurements obtained by Hydro-Quebec, representing common configurations for 120 and 315kV lines. The simulations aimed to replicate typical inspection, exit, and evacuation operations in each environment, with variations in the drone and conductor positions. A total of 36 tests were conducted, and we present 3 non-trivial trajectory experiments in this paper. More results and an online demo can be found at `edu.louispetit.be`.

The simulations showed that the computation time was around 0.01s for the largest environment, i.e., a 30x20m map with 0.3m resolution for a total of 2k 2D cells and 20k nodes in the graph. Our planner is therefore capable of running in real time for frequent trajectory recalculation, e.g., at 50Hz, thanks to (a) the simplification of the environment into a vertical plane of interest, (b) the discretization of the problem, and (c) the integration of a precomputed cost map.

To assess the performance of our algorithm, 3 naive approaches were chosen for comparison: (a) A* for inspection or Dijkstra for exit and evacuation (guaranteed shortest path), (b) Voronoi (based on an A* or Dijkstra search in the Voronoi graph, equidistant from obstacles), (c) Envelope (safest path, based on an A* or Dijkstra search in the outer boundary). These approaches each provide trajectories that would be interesting to perform depending on the situation: the fastest in low battery conditions, the safest in gusty winds, or a compromise between the two. The paths found by these approaches are shown in the left-hand column of Fig. 6. To evaluate the sensitivity of our algorithm to the coefficients in Eq. 1, they have each been varied between 0 and 1 with a step size of 0.05. The results are shown in the right-hand column of Fig. 6. Qualitatively, the paths appears to be similar to the benchmark approaches, depending on the coefficients chosen.

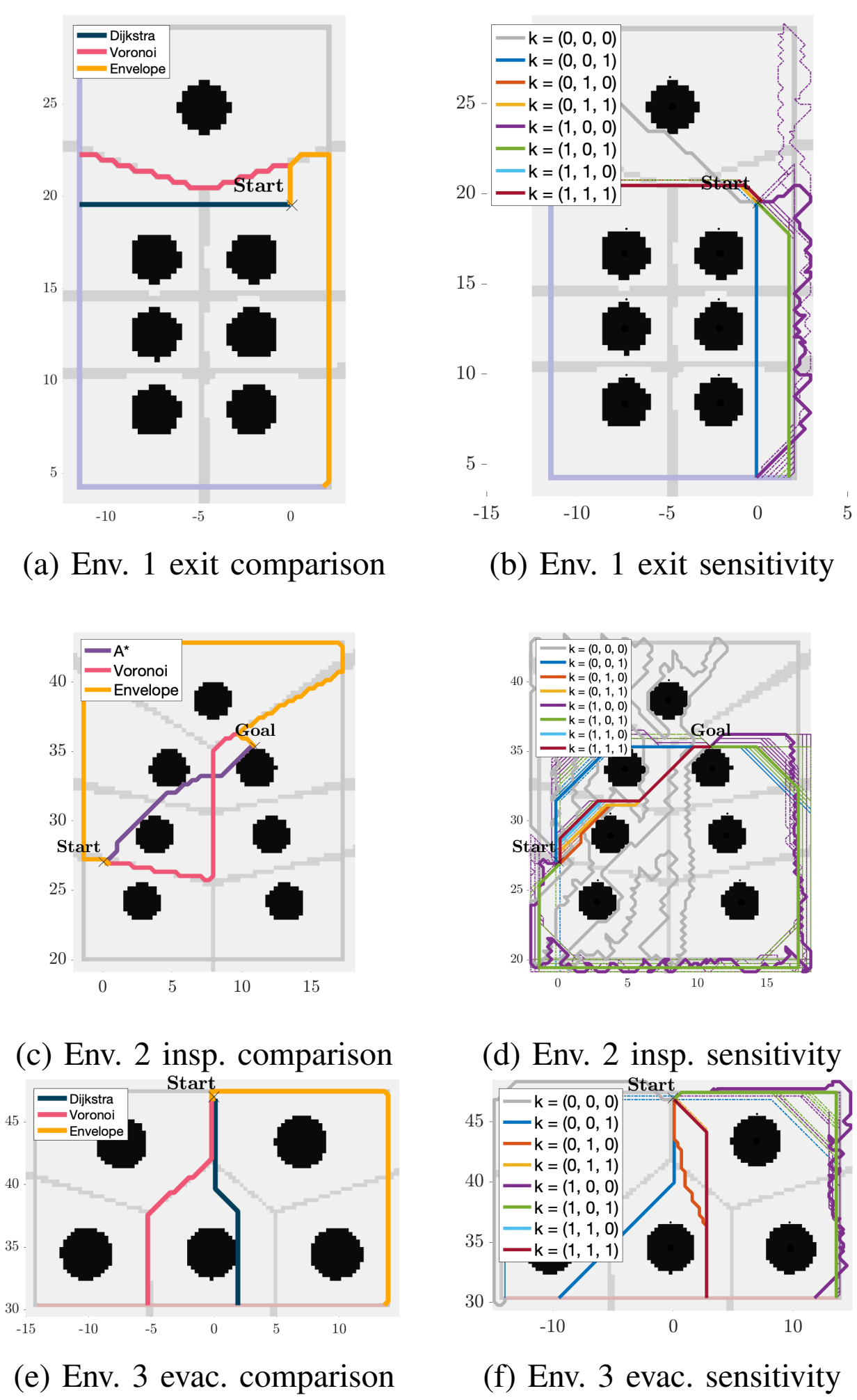

(a) Env. 1 exit comparison (b) Env. 1 exit sensitivity

(c) Env. 2 insp. comparison (d) Env. 2 insp. sensitivity

(e) Env. 3 evac. comparison (f) Env. 3 evac. sensitivity

Fig. 6: Path comparison and sensitivity analysis. The occupied space is in black, the free space in light grey, the Voronoi diagram and grid edges in dark grey, exit edges in blue, and evacuation edges in red. Solid trajectories represent integer coefficients combinations and dotted trajectories are colored according to the closest integer combination in terms of Euclidean distance of coefficients.

For example, trajectories in orange ($k_T$ = 1) are close to A*, and those in green ($k_S$ = 1) are close to the envelope.

In order to analyze results quantitatively, a set of 4 metrics has been defined to assess trajectories: time-averaged distance to closest obstacle, trajectory distance and duration, and total velocity change. It should be noted that some trajectory metrics (travel time, velocity changes) are not comparable with the 3 benchmark approaches, as velocities are not defined in these path planners. These metrics are shown in Fig. 7 for Env. 2 inspection path, as the pattern of results is similar for the other flight missions. The coefficients have been normalized, since only their relative value has an influence on the cost to be minimized. This allows the results to be represented on a 2D plane, since only 2 of the 3 coefficients are independent. The $k_E$ axis is therefore positioned at 45 degrees in the figures. The data with $k_i =$ $(0, 0, 0)$ (grey path in Fig. 6d) is not shown, as it can not be normalized. To ensure figure readability, a color saturation value is set at the 99.5-percentile when the maximum point is more than 1.5 times the range between the minimum point and the saturation value. Outliers above saturation are shown in red and the maximum value is indicated in the figures.

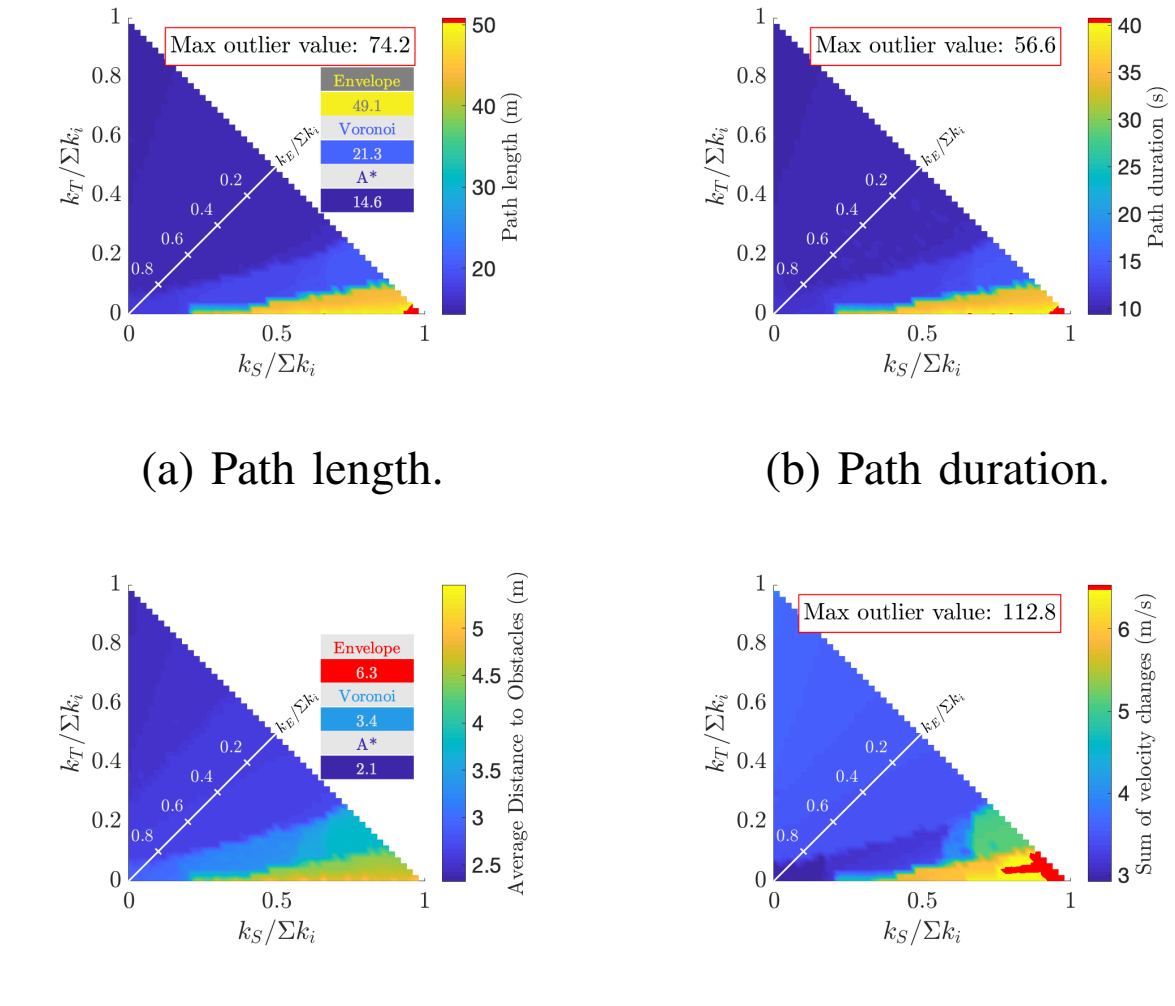

(a) Path length. (b) Path duration.

(c) Avg. distance to obstacles. (d) Total velocity changes.

Fig. 7: Trajectory metrics comparison in Env. 2.

The results confirm the influence of the coefficients: $k_S$ and $k_T$ modulate the trade-off between travel time and distance to obstacles, while $k_E$ reduces velocity changes. One should note that $k_S$ and $k_T$ also have a major influence on velocity changes, which are dependent on travel distance and obstacle clearance. A good set of cost coefficients in Eq. 9 is around the middle point $k_{S,0} = 0.33$, $k_{T,0} = 0.33$, $k_{E,0} = 0.33$. This allows to stay far enough from obstacles while still being fast enough and with few velocity changes. That said, most of the result surface varies little, except for the extreme points close to $k_{S,0} = 1$. One could therefore choose another set of coefficients that would give similar results, especially in the case of an inspection where the goal is unique—unlike an evacuation to a zone—as demonstrated by the similarity of most of the paths in Fig. 6d.

Fig. 7a and Fig. 7c also confirm that varying the coefficients allows moving through the entire metrics range occupied by the benchmark planners. In fact, path length varies from around 150% of the range of benchmark algorithms, and 76% of the range of obstacle clearance. By varying the cost function coefficients, one could make a gradual transition between different desirable behaviors. The choice between these behaviors should depend on risk levels, as explained in Section IV-C. The next subsection validates the adaptability of our method to risk levels.

## B. Risk Adaptability

The risk indices for battery ($\mathcal{BR}$), wind ($\mathcal{WR}$) and GPS ($\mathcal{LR}$) have each been varied from 0 to 1, with an increment of 0.2. Communication risk ($\mathcal{CR}$) has not been varied to reduce the number of possible combinations, but it has the same influence as GPS in Eq. 9. The results are reported in Fig. 8 for an inspection in Env. 1. The desired behavior is observed in Fig. 8a: (a) the yellow trajectory passes outside the envelope when the risk of wind and GPS are high, (b) the cyan trajectory is short and passes further away from the obstacles when the risk of battery and wind are high, (c) intermediate behaviors with other risk levels. Fig. 8b and Fig. 8c show average obstacle distance and path length as a function of wind risk when other risk levels are set to zero. By comparing the values to the comparison algorithms, we observe that our planner switches around 0.5 from A*-like behavior to Voronoi-like behavior. This observation is applicable to other environments, missions and risk factors. The MOAR data points occupy a range of around 90% between A* and Voronoi in terms of clearance distance, and 88% in terms of path length. It confirms that our algorithm is capable of adapting in real time to the level of risk, with an appropriate weighting of navigation objectives.

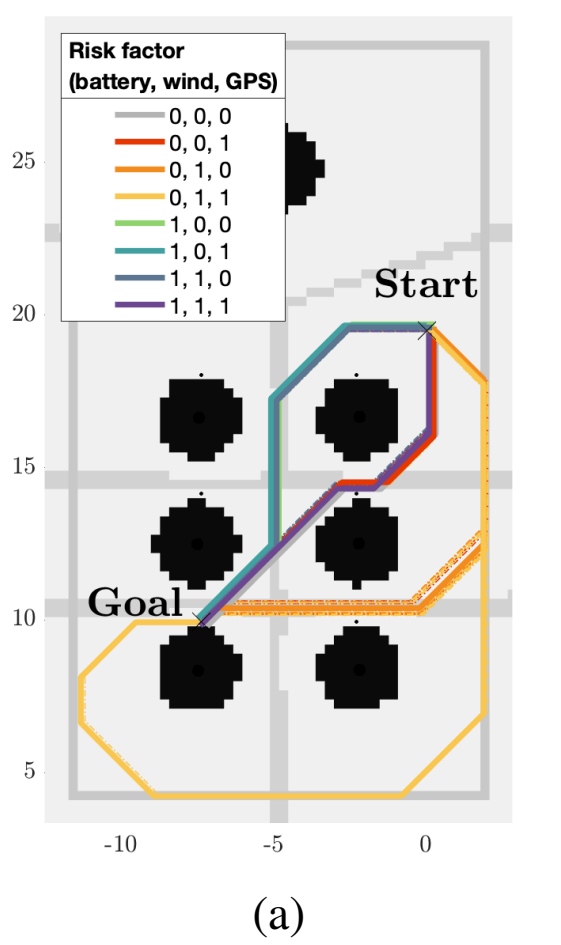

(a)

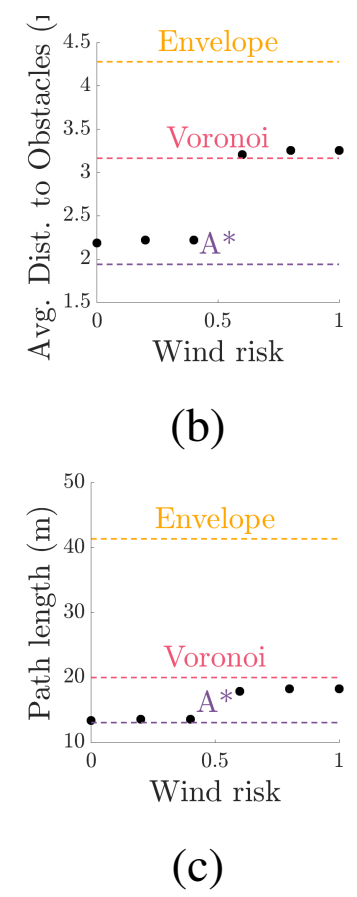

(b)

(c)

Fig. 8: Sensitivity to risk factors: (a) Inspection Path, (b) Average Distance to Obstacles, (c) Path Length. Solid trajectories represent integer factor combinations and dotted trajectories are colored according to the closest integer combination in terms of Euclidean distance of risk indices.

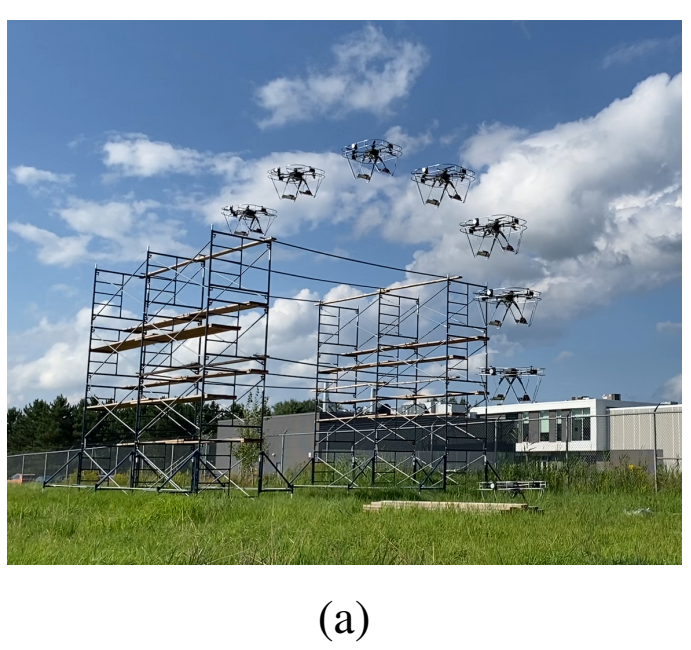

(a)

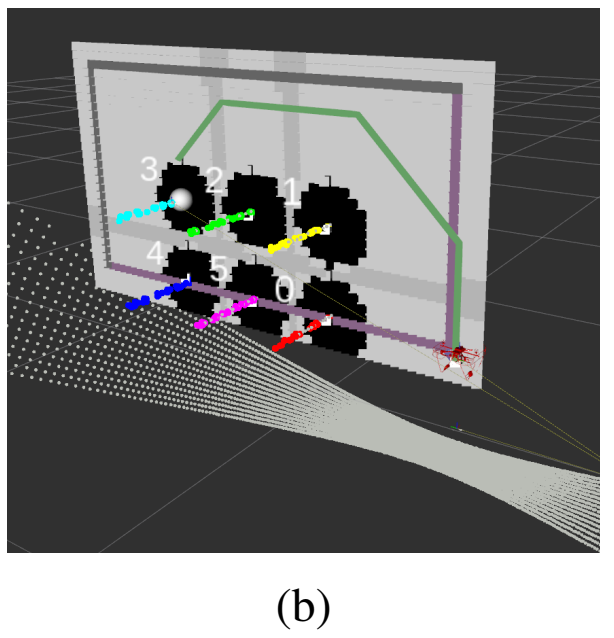

(b)

Fig. 9: Flight test with the LineDrone on a power line model. (a) Path reconstructed from a video. (b) Planned inspection trajectory towards the top left conductor, with $k_{S,0} = 0.43$, $k_{T,0} = 0.14$, $k_{E,0} = 0.43$.

### C. Physical Experiment

The practical feasibility and real-time capabilities were validated through flight tests. Our method was used by the LineDrone for an inspection mission on a power line model with speeds defined between 0.4 and 0.7 m/s. Fig. 9 shows the results of this flight. The UAV covered 17.7m in 36.2s (16.8m and 38.3s planned) with a minimum clearance of 1.8m and an average of 3.7m (1.7 and 3.8 planned). The calculation time on board a Jetson Xavier NX was 9ms. For reasons of operation safety, $k_{T,0}$ has been set low compared with $k_{S,0}$, and the minimum altitude of the navigable space has been defined at 2 meters above ground level. The results show that the drone follows a trajectory that maintains good clearance from obstacles at all times, and makes few changes of direction, while keeping the path fairly short.

## VI. CONCLUSION

In this paper, we propose a novel approach to address the MOPP problem for inspection missions with a UAV. Our method combines a risk-aware cost function, a graph search algorithm, and coefficient adjustment to dynamically adapt to changing risk factors and transition smoothly between different navigation modes. Our contributions to the field of MOPP are as follows: (a) we present a novel framework that integrates risk factors into the cost function, offering precise control over mission objectives and safety considerations, (b) we define a new cost function and cost integration approach, leveraging pre-computed maps to enable real-time computation, (c) we introduce the concepts of damage and insertion costs, which account for potential crash consequences and safe insertion into obstacle zones, (d) we use an adaptive speed planning framework, enabling diverse velocity profiles, (e) we adopt a discrete graph representation for state and action spaces with motion primitives, facilitating exhaustive search for optimal solutions.

Our method is evaluated through extensive simulations and flight tests. We demonstrate the adaptability to changing risk levels, ensuring optimal performance under various operational conditions, e.g., by producing trajectories spanning around 90% of the desired trajectories in terms of length and clearance. In future studies we would like to incorporate data- or model-driven energy consumption computation into the cost function for specific platforms. Future work also include the integration of acceleration profiles for faster vehicles, extending the search space to velocities below cruising speed. Finally, while already applicable in various 3D environments, our approach can be extended to larger spaces that pose real-time computation challenges. To address this, we will explore continuous path representations, such as NURBS [29], and assess alternative solvers, such as genetic algorithms [28].

The adaptability of our method makes it a valuable tool for a wide range of critical missions, from infrastructure inspection to emergency response operations. By addressing the complex interplay of safety, efficiency, and adaptability, our approach contributes to the field of autonomous navigation, enabling robots to navigate challenging environments with a higher degree of autonomy and reliability.